\documentstyle[colap,lexample]{article}
\title{\vspace{-0.5in} Mental State Adjectives: the Perspective of Generative Lexicon}
\author{Pierrette Bouillon \\ 
ISSCO, University of Geneva\\ 
54, route des Acacias\\
1227 Geneva, Switzerland\\ 
{\tt pb@divsun.unige.ch}}

\makeatletter
\long\def\nnfoottext#1{\insert\footins{\footnotesize
    \interlinepenalty\interfootnotelinepenalty
    \splittopskip\footnotesep
    \splitmaxdepth \dp\strutbox \floatingpenalty \@MM
    \hsize\columnwidth \@parboxrestore
    \edef\@thefnmark{}
    {\rule{\z@}{\footnotesep}\ignorespaces\parindent 1em\noindent#1\strut}}}
\long\def\@makefntext#1{\parindent 1em\noindent
                        \hbox to 1.0em{\hss$^{\@thefnmark}$}\ #1
}
\makeatother

\begin{document}

\maketitle
\vspace{-0.5in}
\begin{abstract}
This paper focusses on mental state adjectives and offers a unified analysis
in the theory of {\it Generative Lexicon} (Pustejovsky, 1991, 1995).  We show
that, instead of enumerating the various syntactic constructions they enter
into, with the different senses which arise, it is possible to give them a
rich typed semantic representation which will explain both their semantic and
syntactic polymorphism.

\end{abstract}

\section{Introduction}

Recently\nnfoottext{We would like to thank James Pustejovsky for
extensive discussions on the data presented in this article.  Thanks
also to Laurence Danlos and Graham Russell for their comments.}, work in
computational semantics and lexical semantics has made an interesting
shift.  Motivated by a concern for lexical organization and global
coherence in the structure of lexicon, some researchers have moved
towards more expressive semantic descriptions, as well as more powerful
methods of combining them (see for example Pustejovsky, 1991, 1995;
Briscoe, 1993).

This article will exploit one of these theories, {\it The Generative
Lexicon} (GL: Pustejovsky, 1995), and extend it for the treatment of
French mental adjectives. The following section summarizes the
problematic behaviour of these adjectives. The GL approach is then
described, and a GL analysis of the data.

\section{The Data}

Mental adjectives which denote an {\bf emotional state} or a {\bf
competence} ({\bf agent-oriented}, following Ernst, 1983) present
interesting syntactic and semantic polymorphic behaviour, as noted in
the literature (see for example Lehrer, 1990 and Croft, 1984).  In this
paper, we focus on the representative members of these classes in I and
II:

\begin{itemize}
\item[I] 
{\bf Emotional adjectives}: {\it triste\/} `sad', {\it furieux\/} `angry,
furious', {\it irrit\'e\/} `irritated', {\it heureux\/} `happy', {\it
ennuy\'e\/} `bored'.
\item[II] 
{\bf Agent-oriented adjectives:} {\it intelligent, ing\'enieux\/}
`clever', {\it habile\/} `skilful', {\it adroit\/} `dextrous'.
\end{itemize}

\noindent
Both classes of adjectives exhibit the property of {\bf syntactic
polyvalency}, being able to appear in several distinct contexts, with
optional complement structures (as illustrated in (1), (2) and (3)).  In
the case of {\bf agent-oriented} adjectives, the complement expresses
the {\bf manifestation} of the state and can be realized as an
infinitive with {\it \`a/pour} or {\it de} (examples (2a,b)) or a
prepositional phrase (2c):  (2a,b,c) means that {\it somebody is skilful
in what he does} or {\it how he does it} (see Croft, 1984).

\begin{el}
\ex
Cet homme est triste/habile/furieux\\
``This man is sad/clever/angry{\footnote{We give in quotes a literal 
translation of the French examples.}}''  
\ex
\begin{eli}
\exi
Cet homme est habile de partir\\ 
``This man is skilful to leave''
\exi
Cet homme est habile \`a/pour tricher\\
``This man is skilful at cheating''
\exi
Cet homme est habile au bridge\\
``This man is skilful at bridge''
\end{eli}
\end{el}

In the case of {\bf emotional state} adjectives, this complement
typically expresses the cause of the emotional state and may be realized
as an {\it de-}infinitive or a {\it que-}sentence:  (3a,b), for example,
means that {\it somebody is sad/angry because of something}.  Notice
however that, in some contexts, the complement can also refer to the {\bf
manifestation} of the state, as for {\bf agent-oriented} adjectives
(3c).

\begin{el}
\ex
\begin{eli}
\exi
Cet homme est triste/furieux de partir\\
``This man is sad/angry to leave''
\exi
Cet homme est triste/furieux que tu partes\\
``This man is sad/angry that you are leaving''
\exi
Cet homme est triste en jouant au piano\\
``This man is sad at playing piano''
\end{eli}
\end{el}

Another property exhibited by these adjectives is that of {\bf multiple
semantic selection}:  that is, they are able to predicate of different
semantic types (examples (4) to (7)), namely nouns denoting {\bf
individuals} (the `a' examples), {\bf objects} (b) and {\bf events} (c).
This can however not be generalized to the whole class of mental states
adjectives, as shown by (7), for example.

\begin{el}
\ex
\begin{eli}
\exi
Un homme triste ``A sad man''
\exi
Un livre triste ``A sad book''
\exi
Un examen triste ``A sad exam''
\end{eli}
\ex
\begin{eli}
\exi
Un homme furieux ``An angry man''
\exi
Un livre furieux ``A furious book''
\exi
Une destruction furieuse ``A furious destruction''
\end{eli}
\ex
\begin{eli}
\exi
Un homme ing\'enieux
``A clever man''
\exi
Un livre ing\'enieux
``An clever book''
\exi
Un examen ing\'enieux
``An clever exam'' 
\end{eli}
\ex
\begin{eli}
\exi
Un homme irrit\'e/ennuy\'e\\
``An irritated/bored man''
\exi
\ungram Un livre irrit\'e/ennuy\'e\\
``An irritated/bored book''
\exi
\ungram Une destruction irrit\'ee/ennuy\'e\\
``An irritated/bored destruction'' 
\end{eli}
\end{el}

Finally, the third interesting property manifested by these
adjectives is their pattern of {\bf polysemy}. They exhibit 
different senses depending on the semantic type of the item modified: 
when they predicate of an individual, they normally denote the mental 
state of this individual (8) (but see example (12)). 

\begin{el}
\ex
Un homme triste/ing\'{e}nieux/furieux\\
``A sad/clever/angry man''\\
$\rightarrow$ which is in a sad/clever/angry state
\end{el}

When they modify an event or an object, they can take either 
a {\bf causative} (9b) or a {\bf manifestation} sense 
(9c, 10c and 11c). In the former case, the object or
event is the cause of the state, while in the latter it is the
manifestation of the state.
In some specific contexts, the causative sense is also possible with 
individuals (12).

\begin{el}
\ex
Un livre/voyage triste\\ 
``A sad book/travel''
\begin{eli}
\exi
$\rightarrow$ *which is in a state of sadness
\exi
$\rightarrow$ which causes somebody to be sad
\exi
$\rightarrow$ which is a manifestation of somebody's sadness
\end{eli}

\ex
Un livre/voyage ing\'enieux\\ 
``A clever book/journey''
\begin{eli}
\exi
$\rightarrow$ *which is in a state of cleverness
\exi
$\rightarrow$ *which causes somebody to be clever
\exi
$\rightarrow$ which is a manifestation of the somebody's cleverness
\end{eli}
\ex
Un(e) livre/destruction furieux(se)\\ 
``A furious book/destruction''
\begin{eli}
\exi
$\rightarrow$ *which is in a state of anger
\exi
$\rightarrow$ *which causes somebody to be angry
\exi
$\rightarrow$ which is a manifestation of somebody's anger
\end{eli}
\ex
Un homme triste \`a voir\\ 
``A sad man to see''
\begin{eli}
\exi
$\rightarrow$ *who is in a state of sadness
\exi
$\rightarrow$ the sight of whom causes somebody to be sad
\end{eli}
\end{el}

A complication arises with respect to the polysemous
behaviour of emotion adjectives, in that when they modify an object or
an event (9, 10), they can have both the causal and the manifestation
senses (9b,c).  For some emotion adjectives as {\it furieux} (11),
the manifestation sense is even the only one available (11c).

The remainder of this paper will present an explanation of the syntactic 
and semantic behaviour of these adjectives within the framework of 
Generative Lexicon theory (henceforth GL). In particular, instead of 
enumerating all syntactic constructions and the different senses for 
these adjectives, we will provide a rich typed semantic representation
which explains both the semantic and the syntactic polymorphism 
associated with these classes. This representation and the way to project it
at the syntax level will be the focus of the following section. 

\section{Mental Adjectives in Generative Lexicon}

\subsection{General approach}

In the rest of the article, we will propose the following approach:

\begin{itemize}
\item[(a)] 

to distinguish {\bf emotion adjectives} and {\bf agent-oriented
adjectives} by means of their qualia structure;

\item[(b)] 

to represent the semantic ambiguity of mental adjectives by use of {\bf
dotted types\/} (Pustejovsky 1995, chapter 6.2);

\item[(c)] 
to explain specific semantic selection by the notion of {\bf headedness}
(Pustejovsky 1995, chapter 5.3).

\end{itemize}
The two first points will be the object of section 3.2 and the third
one of 3.3. Section 4 will then focus on emotion adjectives.

\subsection{Two kinds of adjective with dotted type}

The {\bf emotional state} (I) and {\bf agent-oriented} (II) adjectives
will be given the GL representations (13) and (14), respectively.

{\obeyspaces\gdef {\ }}
\global\newbox\codebox
\global\newbox\savedcodebox
\gdef\sverbatim{\bgroup\def\endsverbatim{\egroup\egroup\egroup\mbox{\box\codebo\
x}}\def\savecode{\egroup\egroup\egroup\global\setbox\savedcodebox\copy\codebox}\
\def\par{\egroup\vspace{-0.3em}\hbox\bgroup}\tt\obeylines\obeyspaces\global\set\
box\codebox\vbox\bgroup\hbox\bgroup}
\gdef\savedcode{\copy\savedcodebox}

\newcommand{\avm}[1]{{\setlength{\arraycolsep}{0.8mm}
                       \renewcommand{\arraystretch}{1.2}
                       \left[
                       \begin{array}{l}
                       \\[-4mm] #1 \\[-4mm] \\
                       \end{array}
                       \right]
                    }}
\newcommand{\avmplus}[1]{{\setlength{\arraycolsep}{0.8mm}
                       \renewcommand{\arraystretch}{1.2}
                       \left[
                       \begin{array}{l}
                       \\[-2mm] #1 \\[-2mm] \\
                       \end{array}
                       \right]
                    }}
\newcommand{\att}[1]{{\mbox{\small {\bf #1}}}}
\newcommand{\attval}[2]{{\mbox{\small {\sc #1}}\,=\,{{#2}}}}
\newcommand{\attvalterm}[2]{{\mbox{\small {\sc #1}}\,=\,{\myvalue{#2}}}}
\newcommand{\attvaltyp}[2]{{\mbox{\small{\sc #1}}\,=\,{\myvaluebold{#2}}}}
\newcommand{\myvalue}[1]{{\mbox{\normalsize {\it #1}}}}
\newcommand{\myvaluebold}[1]{{\mbox{\small {\bf #1}}}}
\newcommand{\ind}[1]{{\setlength{\fboxsep}{0.5mm}\,\:\,\fbox{{\tiny #1}} \:}}

\def\implies{\Rightarrow}


\newpage
\refstepcounter{cex}
\medskip\noindent(13)\par
{\centering\tiny
$\avmplus{\att{emotion\_adj}\\
        \attval{eventstr}{\avmplus{
             \attvaltyp{E1}{\tt e1:state}\\
             \attvaltyp{D\_E1}{\tt e2:experiencing\_ev}\\ 
             \attvaltyp{D\_E2}{\tt e3:intellec-act\_ev}\\ 
             \attvaltyp{Restr}{\tt e2 < e1 < e3} }}\\
        \attval{argstr}{\avmplus{
             \attvaltyp{arg1}{\tt x:human}\\
             \attvaltyp{D\_arg1}{\tt e2/e3} }}\\
          \attval{qualia}{\avmplus{\att{(e1.e2).(e1.e3)\_lcp}\\
\attvaltyp{formal}{Adj(e1,x)}\\
\attvaltyp{telic}{P(e3,x,...)}\\
\attvaltyp{agentive}{P(e2,x,...)}
}}
}$

}

\refstepcounter{cex}
\medskip\noindent(14)\par
{\centering\tiny
$\avmplus{\att{agent-oriented\_adj}\\
        \attval{eventstr}{\avmplus{
             \attvaltyp{E1}{\tt e1:state}\\
             \attvaltyp{D\_E1}{\tt e3:intellec-act\_ev}\\ 
             \attvaltyp{Restr}{\tt e1 < e3} }}\\
        \attval{argstr}{\avmplus{
             \attvaltyp{arg1}{\tt x:human}\\
             \attvaltyp{D\_arg1}{\tt e3} }}\\
          \attval{qualia}{\avmplus{\att{e1.e3\_lcp}\\
\attvaltyp{formal}{Adj(e1,x)}\\
\attvaltyp{telic}{P(e3,x,y)}
}}
}$

}
\medskip\noindent 
These structures encode several different aspects of the semantics for these
adjectives.

The {\bf event structure} ({\sc eventstr}) indicates that mental adjectives 
have a complex event structure. They denote mental {\bf state} ({\it e1}) 
(examples (1) to (3)), but they are also able to make reference to 
{\bf events}, the cause of the state ({\it e2}) and/or its 
manifestation ({\it e3}) (as shown in examples (4) to (7)). 
The {\it Restr(iction)} relation indicates the temporal precedence between 
the state and the two events: the cause ({\it e2}) must precede the 
state and the manifestation ({\it e3}) must follow it. The two events are 
default events, as the adjective remains a state, even when it has a 
causative sense, contrary to real transitions (accomplishment or 
achievement), like {\it couler} `sink', for example (as pointed out in 
Pustejovsky, 1995, chapter 10)

The {\bf argument structure} ({\sc argstr}) specifies that mental 
adjectives select for two arguments, one for {\bf human} ({\it arg1}) and a 
second for {\bf event} (see Croft, 1984, for a similar view). 
The second is a default argument ({\it D\_arg1}) as it need not to be 
present at the syntactic level (as shown in examples (1)). As 
{\bf agent-oriented adjectives} refer to the manifestation 
of the state (examples (2)), the second argument is {\it e3}, the event
which follows the state. It is subtyped as an intellectual act. 
For {\bf emotion adjectives}, the second argument is {\it e2} or 
{\it e3}, as they can refer either to the manifestation of the state 
(example (3c)) or its cause (examples (3a,b)). {\it e2} is subtyped as 
an experiencing event, as we consider that the cause of an emotion 
corresponds to the experiencing of something. Following Croft (1990), 
we think that there are two processes implied in a causal emotional 
state: an experiencer must direct his or her attention to a stimulus 
and this causes the experiencer to enter in a mental state.  

The {\bf qualia structure} ({\sc qualia}) encodes the basic semantic 
type of a word (its Lexical Conceptual Paradigm, or {\it LCP\/}) and specifies 
how it is linked to other events and arguments 
of the event and argument structures (see Pustejovsky, 1995, chapter 6).
To do this, it can use four possible different roles: the {\sc formal} role 
encoding the basic semantic type(s) of the word, the {\sc constitutive} role 
its constitutive elements, the {\sc telic} role its purpose or function and 
the {\sc agentive} role the factors involved in bringing it about. In terms 
of temporal relations, the qualia encode
specific constraints on the relative temporal ordering of the
values of the qualia. That is, the event involved in the {\sc
agentive} role precedes that state existing in the {\sc formal}, and
the associated {\sc constitutive} value, should there be
one. Finally, the {\sc telic} role is inherently a temporal consequence
of the {\sc formal}, cf.\ (15).

\begin{el}
\ex
{\sc agentive} $\leq$ {\sc formal} and {\sc const} $\leq$ {\sc telic}
\end{el}

In the case of mental adjectives, the qualia in (13) and (14) makes
explicit that they denote a complex or {\bf dotted} type (written {\it
type.type}), which is the product of basic types, {\it e1} and {\it e3}
for {\bf agent-oriented} adjectives and {\it e1}, {\it e2} and {\it e3}
for {\bf emotive} ones.  Each of these types can be projected
independently, if no other constraints apply (see 3.3).  The {\bf state}
{\it e1} is encoded in the {\sc formal}; an {\bf event} encoded
in the {\sc agentive} role ({\it e2}) denotes the {\bf cause} or origin of the
state, i.e.\ the experiencing event; encoded in
the {\sc telic} role ({\it e3}), it denotes then the {\bf manifestation}
of the state, i.e.\ the intellectual act.

In other words, the GL representation for 
{\bf emotion} adjectives (13) stipulates that somebody ({\it x}) is 
in a state because of an experiencing event ({\it e2}), which can have a 
further manifestation ({\it e3});\footnote{For a similar view see Anscombres 
(1995) who distinguishes internal feeling and 
external attitude. He considers then that a feeling can have a external 
manifestation.} that for {\bf agent-oriented} adjectives (14) 
specifies that somebody is in a state which can have a manifestation.

\subsection{The notion of head}

However, not all mental adjectives will be able to project the two types they
denote (i.e.\ state and event), depending on the event headedness.  In GL, the
notion of {\em head} provides a way of indicating a type of foregrounding and
backgrounding of event arguments. In doing this, it specifies how to project
the qualia representation and acts as a filter to constrain the set of
projectable qualia: the headed event projects the formula associated with that
event and it is this formula which needs to be saturated at the syntax level
(Pustejovsky, 1995, Chapter 6.2.5).

For mental adjectives, two kinds of headedness are possible. The adjective can
be headed either on the state or the event it denotes. Moreover, some
adjectives will be unspecified regarding the head and will therefore be able
to be headed on any of the subevents of the event structure.  In the
following, we will first focus on the two different kinds of headedness,
applying to the state or one of the events, and then show the consequences of
an headless structure.

\subsubsection{Event structure headed on the state}

\noindent
The adjective is projected via the template {\it P(e1,x)} in the formal role.
It therefore denotes the mental state of an individual (16a, 17a) and 
requires only one argument {\it x} of type {\it human}. Complements are 
however possible if they make direct reference to the agentive (as in (16b,c), 
where the complement is the cause of the emotional state) or telic roles 
(as in (17b,c), where it is the manifestation).

\begin{el}
\ex
\begin{eli}
\exi
Je suis triste/furieux ``I'm sad/furious''
\exi
Je suis triste/furieux de partir \\
``I'm sad/furious at leaving''
\exi
Je suis triste/furieux que tu partes \\
``I'm sad/furious that you are leaving''
\end{eli}
\ex
\begin{eli}
\exi
Je suis ing\'enieux ``I'm clever''
\exi
Je suis ing\'enieux aux \'echecs \\
``I'm clever at playing chess''
\exi
Je suis ing\'enieux de partir \\
``I'm clever to leave''
\end{eli}
\end{el}

The qualia representation is rich enough to explain the syntactic
{\bf polyvalency} shown in (16) and (17). There are indeed two ways of 
referring to a quale role: 

\paragraph{Direct saturation of a quale role.}
The complement is identified as a subtype of the experiencing event or the
intellectual act.  In (16b) and (17b), for example, the complement directly
saturates the event {\it e2} or {\it e3} (as the qualia structures in (18) and
(19) make explicit).  {\it Partir} is indeed a subtype of the experiencing
event sort ({\it partir $<$ experiencing\_event}) and {\it les \'echecs}
(chess) of the intellectual act one ({\it les \'echecs $<$
creative-intellectual\_act}).
By contrast, in order for (20) to be an acceptable sentence, {\it
\^{e}tre malade} `be ill' must be reconstructed, non-standardly, as an
intellectual act.

\refstepcounter{cex}
\medskip\noindent(18)\par
{\centering\tiny
$\avmplus{\att{}\\
\attvaltyp{formal}{triste(e1,je)}\\
\attvaltyp{telic}{P(e3,je)}\\
\attvaltyp{agentive}{partir(e2,je)}
}$\par}

\refstepcounter{cex}
\medskip\noindent(19)\par
{\centering\tiny
$\avmplus{\att{}\\
\attvaltyp{formal}{ing\'enieux(e1,je)}\\
\attvaltyp{telic}{jouer(e3,je,\'echecs)}
}$\par}

\begin{el}
\ex\ungram
Je suis habile \`a \^etre malade\\ 
``I'm skilful at being ill''
\end{el}

\paragraph{Saturation of the object of the experiencing or intellectual
act event.}

In (16c) and (17c), the complement is the object ({\it y}) of an implicit 
event and the saturation of the quale is only possible because the complement 
can be coerced to the type expected for the complement (the experiencing event
or intellectual act): (16c) means that {\it I'm sad/furious because I 
experience your leaving} (as (21) makes explicit) and (17c) 
that {\it I'm ingenious at performing the intellectual act whose object is the
departure} (as in (22)). There is no further specification available for the
{\it exp\_ev} or {\it intellectual-act\_ev} variable.

\refstepcounter{cex}
\medskip\noindent(21)\par
{\centering\tiny
$\avmplus{\att{}\\
\attvaltyp{formal}{triste(e1,je)}\\
\attvaltyp{telic}{P(e3,je)}\\
\attvaltyp{agentive}{exp\_ev(e2,je,que-tu-partes)}
}$\par} 

\refstepcounter{cex}
\medskip\noindent(22)\par
{\centering\tiny
$\avmplus{\att{}\\
\attvaltyp{formal}{ing\'enieux(e1,je)}\\
\attvaltyp{telic}{intellec-act\_ev(e3,je,partir)}
}$\par}
\medskip

These two ways of saturating a quale explain what Croft (Croft, 1984) and
Ernst (1985) have called the verbal/factive ambiguity of two arguments
agent-oriented adjectives (see also Kiparsky and Kiparsky, 1979). When the
event is saturated, we get the eventual sense: in (17b), cleverness is
predicated of the manner of playing chess (structure 19); when the object of
the event is saturated, we get the factive sense so that in (17c) cleverness
is predicated of the fact of leaving (structure 22).

\subsubsection{Event structure headed on an event}
\noindent
Recall that the adjective denotes one or two events,  i.e.\ {\it e2} or {\it
e3} in (13) and (14).
When the event structure is headed on one of these, the adjective is 
projected via the agentive or the telic role, i.e.\ the template 
$P(e2,x,\ldots)$ or $P(e3,x,\ldots)$.  It therefore selects for an event 
and gets the causative or manifestation sense (examples (23) and (24)). 
However, that does not mean that the noun must be an event, but only 
that its semantic representation, or general knowledge concerning its semantic
type, should provide an event, as shown in the next examples (23) and (24).

\begin{el}
\ex
Ce livre est triste ``The book is sad''
\begin{eli}
\exi
$\rightarrow$ whose {\bf reading} causes somebody to be sad
\exi
$\rightarrow$ whose {\bf writing} causes somebody to be sad
\exi
$\rightarrow$ whose {\bf writing} is the manifestation of somebody's 
sadness 
\end{eli}
\ex
Ce sapin est triste ``The pine tree is sad''
\begin{eli}
\exi
$\rightarrow$ whose {\bf experiencing} causes somebody to be sad
\end{eli}
\end{el}

In (23), the modification by the adjective is possible as {\it livre} (book) 
contains in its qualia structure two events, namely {\it lire} (to read) 
(telic of {\it livre}) and {\it \'ecrire} (to write) (agentive of 
{\it livre}) (see Pustejovsky and Bouillon, 1995, for the qualia 
representation of {\it livre}). Two causative interpretations 
(23a,b) and one manifestation (23c) are therefore possible. Notice that when 
the events are defined in the lexical semantics of the word, the experiencing 
and the manifestation are intentional and controlled 
(the experiencing is {\bf active}, following Lehrer, 1990).

In (24) on the other hand, there is nothing contributed by the {\it tree} 
per se to how the experiencing is achieved (as the noun has no telic nor 
agentive), except for it being a physically manifested object with 
extension. In this case, it is the properties inherited through the 
formal (and not the lexical semantics of the word) that suggests how it 
can be experienced. For this reason, the experiencing is not controlled, 
nor intentional (it is {\bf stative}). The manifestation sense is impossible 
as {\it sapin} (versus {\it book}) has no intellectual act in its qualia.

\subsubsection{Unheaded event structure.}
\noindent
If certain adjectives can be restricted to be headed either on the
event or the state, others can be left underspecified regarding the head. 
In this last case, the adjective can then be projected via the formal or 
the telic/agentive roles and combines the two or three different senses: 
stative, causative and/or manifestation, depending on the number of events
it can refer to (one for agent-oriented adjectives (see (14)), two 
for emotional ones (see (13))). This is the case of the adjectives 
{\it triste} and {\it ing\'{e}nieux} which will get respectively the three and 
two senses, as illustrated in (8) to (10); in (8), {\it triste} and 
{\it ing\'{e}nieux} have the head on the formal role and the adjectives 
have a stative sense; in (9c) and (10c) on the telic: they have a 
manifestation sense. In (9b), {\it triste} has the head on the agentive 
and receives its causative sense. Their semantic polymorphism is then 
explained, without having to list the different senses. 

Remember however that all emotional state adjectives which combine
a stative and a eventual meaning will not be able to get the three meanings: 
the emotion adjective {\it furieux}, for example, cannot have the 
head on the 
agentive, as shown in (11b): {\it un livre furieux} cannot get the causative
meaning. The question is then what prevents this adjective 
from having the head on the agentive role? A first attempt at tackling 
the problem 
follows from the observation that most emotion adjectives ending in 
{\it -eux} (with causal complement and not derived from psychological 
verbs, as {\it ennuyeux}, {\it outrageux}, etc.) behave in the same way 
(see the list in (25)) and that, more generally, the suffix plays a 
crucial role in restricting 
the head (see Anscombres, 1995, for a similar view and section 4 for other
examples of the influence of the suffix). It seems therefore not too 
preliminary to think that the {\it -eux} suffix acts as a filter on the 
head for this kind of adjectives. However, the formal 
representation of the suffixes and the way it interacts with the representation
of the stem remain to be investigated.

\begin{el}
\ex
{\it heureux, anxieux, malheureux, honteux, soucieux}, etc.
\end{el}

In this section, we explained the polyvalency of mental adjectives. 
We are now able to show how the head distinction is relevant 
to classify emotional states adjectives and explain their 
semantic selection.

\section{Semantic selection: the case of emotion adjectives}

On the basis on the headedness configuration, we will distinguish 
three classes of French emotion adjectives, exemplified in (26), (27) 
and (28):

\refstepcounter{cex}
\begin{itemize}
\item[(26)] {\bf Adjectives headed on the state}: 
\item[] {\it f\^ach\'e} `angry', {\it ennuy\'e} `bored', {\it irrit\'e} `irritated', etc. 

\refstepcounter{cex}
\item[(27)] {\bf Adjectives headed on the agentive}: 
\item[] {\it ennuyant} `boring', {\it pr\'eoccupant} `worrying', 
        {\it agr\'eable} `nice', {\it admirable} `wonderful', 
        {\it effroyable} `appalling', etc.

\refstepcounter{cex}
\item[(28)] {\bf Headless adjectives:}
\item[] {\it triste} `sad', {\it heureux} `happy', 
        {\it furieux} `angry', `furious', etc.
\end{itemize}

Those in (26) will have the head on the state and will get only the
stative sense. As predicted, they will not be able to modify an event or
an object, as illustrated in (29).

\refstepcounter{cex}
\begin{itemize}
\item[(29)] *Un livre f\^ach\'e/ennuy\'e/irrit\'e\\
``An angry/bored/irritated book''
\end{itemize}

The ones in (27) will have the head on the agentive and will receive
only a causative sense. As a result, they will keep this causative sense, even
when they modify a noun of type {\it human} (30).

\refstepcounter{cex}
\begin{itemize}
\item[(30)] Un homme ennuyant/pr\'eoccupant/ admirable/effroyable\\
$\rightarrow$ which causes somebody's trouble/\\ anxiety/admiration/fright
\end{itemize}

Finally, those in (28) will not be specified regarding the head (they are 
headless) and will be able to combine the three senses (except when 
the suffix acts as a filter, as shown in (25)). They can therefore
modify nouns of type {\it human}, {\it object} and {\it event} (5) and 
will be ambiguous when they modify a noun of type {\it human}, as an 
human can be either in a mental state or the object of an experiencing 
event (31a,b). In the case of ambiguity, it is striking to see that French 
syntax distinguishes clearly the two senses. In (31a), it is the 
prenominal position of the adjective and in (31c) the choice of the 
preposition {\it \`a} (versus {\it de} as in (5b)) which give rise to 
the causative sense (versus the stative one).

\begin{el}
\ex
\begin{eli}
\exi
De tristes enfants ``Sad children to see''\\
$\rightarrow$ which cause the sadness of the persons which experience them
\exi
Des enfants tristes ``Sad children''\\
$\rightarrow$ which are in a sad state
\exi
Un homme triste \`a voir\\
``A sad man to see''\\
$\rightarrow$ which causes the sadness of the persons which see him
\end{eli}
\end{el}

As a result of this, (32a) (vs.\ (32b)) will be impossible: in (32a), 
the prenominal position of the adjective forces the causative sense, 
giving rise to an incompatibility as two different nouns 
(namely {\it enfant} (children) and {\it mort de leur m\`ere} 
(death of the mother)) try to saturate the same variable {\it y}, i.e.\ 
the object of the experiencing.

\begin{el}
\ex
\begin{eli}
\exi
*De tristes enfants de la mort de leur m\`ere
\exi
Des enfants tristes de la mort de leur m\`ere
\end{eli}
\end{el}

To finish, notice that in (28), (29) and (30), the role of the suffix 
appears clearly: for emotion adjectives, the {\it -\'e} suffix 
constrains the head to be on the state and {\it -ant/-able} on 
the causative event. It also explains possible divergences between 
French and English.   

\section{Conclusion}

In this article, we extended GL to the treatment of French mental state
adjectives. We showed how GL can adequately account for the following:

\begin{itemize}
\item[(a)] {\bf Avoiding the multiplication of entries}. The different senses of the
mental adjectives (examples (11) to (14)) and their polyvalency (examples
(3) to (5)) follows from the qualia representation.

\item[(b)] {\bf Explaining the links between the different senses of mental 
adjectives}. 
The qualia structure we proposed in (15) and (16) makes explicit the 
links between the different senses of mental adjectives 
(mental state of an individual, causative and manifestation). 
In (15), the qualia structure specifies that emotional states are 
caused by a causal event and can have a further manifestation; in (16), 
that the agent-oriented state can have a further manifestation.
\item[(c)]
{\bf Explaining the semantic selection of mental state adjectives}. 
The specific semantic selection of mental adjectives follows from the
headedness system.
\item[(d)]
{\bf Distinguishing two kinds of adjectives}: those which denote simple type
({\it rouge} (red), {\it grand} (big), etc.) and those like mental adjectives
which denote dotted type. This distinction is in accordance with the classical
distinction drawn between stative adjectives and dynamic ones, which,
following Quirk et al., 1994:434, denote qualities that are thought to be
subject to control by possessor. GL allows this distinction to be
characterized and given a more formal representation, an adjective being
dynamic if it refers to the cause or its further manifestation.
\end{itemize}

\end{document}